\documentclass[12pt, letterpaper]{article}
\usepackage{amsmath, longtable, calc, multirow, booktabs, graphicx, enumitem,
pdfpages, catchfile}
\usepackage[margin=1.25in]{geometry}
\usepackage[utf8]{inputenc}
\usepackage[T1]{fontenc}
\usepackage{charter}
\usepackage{CJKutf8}

\usepackage[style=apa, backend=biber]{biblatex}
\usepackage{hyperref}
\hypersetup{colorlinks=true, allcolors=blue}
\usepackage{siunitx}
\usepackage{tabularx}
\newcolumntype{Y}{>{\raggedleft\arraybackslash}X}
\usepackage{fancyhdr}
\newcommand{\shorttitle}[1]{
  \newcommand{\short}{#1}
}
\newcommand{\affiliation}[1]{
  \newcommand{\showaffil}{#1}
}
\newcommand{\contact}[1]{
  \newcommand{\showcontact}{\href{mailto:#1}{#1}}
}
\newcommand{\showextra}{
  {\small
    \begin{center}
      \showaffil{} -- \showcontact
    \end{center}
  }
}

\newcommand{\cat}[1]{\textsc{#1}}
\newcommand{\code}[1]{\texttt{#1}}
\newcommand{\foreign}[1]{\textit{#1}}

\date{}

\usepackage{expex} 
\lingset{everygla=, belowglpreambleskip=0ex, aboveglftskip=0ex}
\usepackage{forest} 

\makeatletter
\catcode`\_=8
\makeatother

\title{A Computational Approach to Analyzing Language Change and Variation in the Constructed Language Toki Pona}
\shorttitle{Language Change and Variation in Toki Pona}
\author{Daniel Huang\textsuperscript{1}, Hyoun-A Joo\textsuperscript{2}}
\affiliation{
  \textsuperscript{1}Georgia Institute of Technology -- \href{mailto:dzh@gatech.edu}{dzh@gatech.edu} \\
\textsuperscript{2}Georgia Institute of Technology}
\contact{joo.hyouna@gatech.edu}
\begin{document}
\maketitle
\showextra 
\thispagestyle{empty} 

\begin{abstract}
  This study explores language change and variation in Toki Pona, a constructed language with approximately 120 core words. Taking a computational and corpus-based approach, the study examines features including fluid word classes and transitivity in order to examine (1) changes in preferences of content words for different syntactic positions over time and (2) variation in usage across different corpora. The results suggest that sociolinguistic factors influence Toki Pona in the same way as natural languages, and that even constructed linguistic systems naturally evolve as communities use them.
\end{abstract}

\section{Introduction}

Human language typically evolves through centuries of usage, adaptation, and cultural exchange, gradually developing vocabulary and changing grammar. Thus, language change is a natural process and studying it ``helps us explain the features of language structure because it provides a window onto how those structures come into being and evolve'' \parencite[p.~1]{bybee2015}. In contrast with these natural languages, constructed languages are artificially created with an intentionally designed phonology, grammar, orthography, and vocabulary. Constructed languages enjoy continued fascination, evidenced by the invention of thousands of different artificial languages in the past few centuries \parencite{peterson2015}. These include languages like Esperanto intended for international communication and artistic creations like Elvish and Klingon that bring fictional worlds to life.

Constructed languages provide a unique environment to study language variation and change when adopted by a speaking community. When a language is deliberately designed with specific constraints or features, we can observe how these structures evolve over time and how speakers innovate within existing frameworks. Constructed languages, therefore, provide a sandbox to identify drivers of change.

Toki Pona, the language examined here, tests the extremes of what a language can be. It was created by Canadian linguist Sonja Lang in 2001, designed with approximately 120 words and a grammar with very few rules. \textcite{lang2014} published \textit{Toki Pona: The Language of Good}, documenting its grammar and core vocabulary. Today, Toki Pona is primarily spoken on online platforms such as Reddit, Discord, and Facebook, with tens of thousands of community members and thousands of conversationally proficient speakers, making it the second most spoken constructed language in the world after Esperanto \parencite{meulen2021}. Because of its small lexicon, easy-to-parse grammar, and large community, Toki Pona is an ideal subject for a computational study on real language use.

\section{Toki Pona: fluid word classes and transitivity}
\label{sec:fluidwordclasses}

Toki Pona is an isolating language and exhibits a strict SVO order. All words, except proper names, are written in lowercase. Nouns do not inflect for number or definiteness, and verbs do not inflect for tense or aspect. Content words in Toki Pona can be used as a noun, adjective, adverb, or verb, i.e., the word classes are fluid. To delimit phrases, Toki Pona uses several particle words. Consider the usage of \foreign{moku} 'to eat' in (\ref{ex:moku}).\footnote{All Toki Pona examples were glossed using the Leipzig Glossing Rules \parencite{leipzigglossingrules}.}

\pex\label{ex:moku}
\a \begingl
\glpreamble \foreign{moku} as a transitive verb //
\gla jan li moku e kili //
\glb person \cat{pred} eat \cat{tr} fruit //
\glft `The person is eating fruit.' //
\endgl \label{ex:moku-e}
\a \begingl
\glpreamble \foreign{moku} as a noun //
\gla moku ni li suwi //
\glb food \cat{dem} \cat{pred} sweet //
\glft `This food is sweet.' //
\endgl \label{ex:moku-n}
\xe

The particle \foreign{li} in (\ref{ex:moku}) marks the following phrase as the predicate, which can be a noun with zero copula, an adjective, or a verb. The particle \foreign{e} in (\ref{ex:moku-e}) marks the following phrase as the direct object of a verb. Thus, when \foreign{moku} is used after \foreign{li}, it takes on the lexical meaning 'to eat.' However, when used at the beginning of the sentence in (\ref{ex:moku-n}), it must be a noun and therefore converts, now meaning 'that which is eaten, food.'

This fluidity extends to transitivity. If supplied with a direct object marked by \foreign{e}, nouns and adjectives are converted into transitive verbs. When an adjective is converted to a transitive verb, it typically becomes causative, as \foreign{pona} 'good' in (\ref{ex:pona-e}). In contrast, in (\ref{ex:li-pona}), \foreign{pona} is used as a predicative adjective without a direct object assigned, describing the subject.

\pex\label{ex:pona}
\a \begingl
\glpreamble \foreign{pona} as a predicative adjective //
\gla soweli li pona //
\glb {land animal} \cat{pred} good //
\glft `The dog is good.' //
\endgl \label{ex:li-pona}
\a \begingl
\glpreamble \foreign{pona} as a transitive verb //
\gla lipu li pona e sona //
\glb document \cat{pred} make.good \cat{tr} knowledge //
\glft `Books make knowledge better.' / `Books improve knowledge.' //
\endgl \label{ex:pona-e}
\xe

This feature of fluidity presents an opportunity to analyze the distribution of content words in the different word classes and examine variation in their use. Because the particles that determine the word class of the content words take specific positions within the syntactic structure of Toki Pona as laid out above, the frequencies of different words in particular syntactic positions can be studied. The following research questions informed the study: (1) how do the frequencies of words in different syntactic positions change over time, and (2) to what extent does the usage of content words in different syntactic positions differ between informal and formal contexts?

\section{Methods}

\subsection{Corpora}

The data came from two corpora. The first, \foreign{ma pona pi toki pona} \parencite{mapona2025}, means ``a good place for Toki Pona,'' and is a Discord server with over 16,000 members. Discord is a realtime chat application that hosts large communities with several parallel chatrooms. As such, it contains primarily informal, conversational data. The downloaded channels contained 5.97 million total sentences, 1.23 million of which were scored as Toki Pona sentences with a total of 6.39 million tokens of Toki Pona, spanning from 2016 through January 2025.

The second corpus, \foreign{poki Lapo} \parencite{lapo2025}, means ``the collection named Lapo,'' and is a monolingual Toki Pona corpus with long-form content like books, poetry, song lyrics, comics, and blog posts. It is continually updated with new published works and spans from 2002 to 2025.
Because the two corpora represent different uses of Toki Pona (more conversational and informal versus more formal), a comparison between them will allow for insights regarding patterns of use and variation.

\subsection{Filtering and tokenizing data}

A significant majority of the messages in the informal corpus, \foreign{ma pona pi toki pona}, is in English and had to be filtered out.

The Python library \code{sona-toki} `language knowledge' by \textcite{danielson2025} takes a text as input, cleans the text to remove irregularities such as duplicated characters and punctuation, splits the text into sentences and further into tokens, and scores each of the sentences for whether or not they are Toki Pona. After splitting the text into tokens, it uses a variety of heuristics such as the ratio of Toki Pona words to non-Toki-Pona words and the phonotactic restrictions for proper names to generate a final output: a filtered list of sentences segmented into tokens.

\textcite{muni2024} uses a specific configuration of \code{sona-toki} to create an n-gram corpus, that is, a dataset mapping Toki Pona phrases of various lengths to their frequencies. The same configuration was used to extract and filter sentences from both corpora in this research.

\subsection{Parsing sentences}

Before tagging parts of speech, each sentence had to be converted into a hierarchical structure in order to account for syntactic ambiguity (Section~\ref{sec:resolving-syntactic-ambiguity}) and more easily address features like transitivity. For instance, the particle \foreign{e} that determines whether a verb is transitive could be separated from the verb by adverbs, but placing the phrases in a hierarchical structure makes the connection more easily identifiable. Therefore, the first author developed a parser for Toki Pona using the \textcite{earley1970} algorithm with a specific implementation in the programming language JavaScript called \textit{nearley} \parencite{nearley}.

The Earley parsing algorithm defines a context-free grammar, a kind of phrase structure grammar \parencite{chomsky1956}, which is a set of rules in the form of $X \rightarrow Y ... Z$. This allows the syntax of a language to be described by breaking down phrases into their constituents with statements like \code{PP -> P NP} (a prepositional phrase can be broken down into a preposition followed by a noun phrase). The first author created simplified phrase structure grammar based on the Toki Pona grammar rules, illustrated in Table~\ref{tab:code1}.

\begin{table}[hbt!]
  \caption{An example of a simplified Toki Pona-like context-free grammar\protect\footnotemark}
  \label{tab:code1}
  \begin{center}
    \begin{tabular}{p{0.25\linewidth} | p{0.65\linewidth}}
      \textbf{Grammar Rule} & \textbf{Description} \\
      \hline
      \code{S -> P$_{1}$ "li" P$_{2}$} & A sentence (S) is composed of phrase (P$_{1}$), the particle \foreign{li}, and phrase (P$_{2}$), where P$_{1}$ is the subject and P$_{2}$ is the predicate. \\
      \code{P -> C} & A phrase (P) can consist of a single content word (C). \\
      \code{P -> P C} & A phrase (P) can alternatively consist of a combination of a phrase and a content word (C), whereby C is an adjective that must be attached to the phrase. \\
    \end{tabular}
  \end{center}
\end{table}
\footnotetext{The full grammar used to process the corpora can be found online at \url{https://github.com/cubedhuang/ilo-nasin-sin/blob/main/src/lib/grammar.ne}, with an interactive demo at \url{https://nasin.nimi.li}.}

\begin{figure}[hbt!]
  \centering
  \caption{The resulting parse of \foreign{jan pona li moku pona} `the good person eats well' generated by \textit{nearley} with the grammar in Table~\ref{tab:code1}.}
  \label{fig:parse1}
  \begin{forest}
    sn edges/.style={for tree={
    parent anchor=south, child anchor=north}},
    sn edges
    [S
      [P
        [P
          [C [\foreign{jan},tier=word]]
        ]
        [C [\foreign{pona},tier=word]]
      ]
      [\foreign{li},tier=word]
      [P
        [P
          [C [\foreign{moku},tier=word]]
        ]
        [C [\foreign{pona},tier=word]]
      ]
    ]
  \end{forest}
\end{figure}
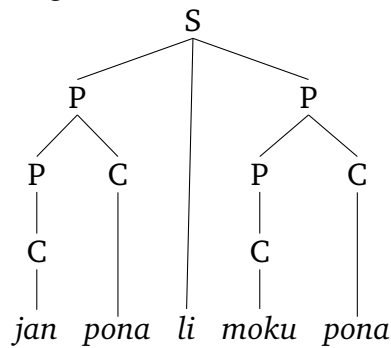

\pagebreak
The \textit{nearley} library will then use the grammar rules in order to generate a program that takes a sequence of tokens as input and outputs a resulting structure (or fails if the tokens do not conform to the grammar). Using the grammar above results in the hierarchical structure in Figure~\ref{fig:parse1}.

\pgfkeys{/pgf/inner sep=0.2em}

This hierarchical structure allows for more advanced algorithm-based part-of-speech tagging by segmenting the sentence into phrases and revealing the structure of those phrases beforehand.

\subsection{Resolving syntactic ambiguity}
\label{sec:resolving-syntactic-ambiguity}

Toki Pona, like natural languages, has syntactic ambiguity. One instance of syntactic ambiguity is in prepositions. Toki Pona's prepositions are a closed set of five words (\foreign{lon} 'at,' \foreign{tawa} 'to,' \foreign{tan} 'from, because of,' \foreign{kepeken} 'by means of, with,' \foreign{sama} 'like'), but all prepositions can also function as content words based on context. This is different from the case of fluid word classes laid out in Section~\ref{sec:fluidwordclasses}. Examples in (\ref{ex:tawa}) illustrate the different uses of \foreign{tawa} 'to; moving.'

\pex\label{ex:tawa}
\a \begingl
\glpreamble \foreign{tawa} as a prepositional predicate //
\gla jan li tawa sike //
\glb person \cat{pred} to circle //
\glft `The person goes to the ball.' //
\endgl \label{ex:tawa-prep}
\a \begingl
\glpreamble \foreign{tawa} as a non-prepositional content word //
\gla jan li tawa sike //
\glb person \cat{pred} move circle //
\glft `The person moves circularly.' / `The person spins.' //
\endgl
\xe

The Earley algorithm is particularly useful for such cases of ambiguity because it generates all possible syntactic interpretations of a string of tokens. However, only one parse should ultimately be chosen for each sentence.

To choose the most probable parse, the first author developed a heuristic scoring algorithm that favors specific patterns over others. For the preposition ambiguity illustrated in (\ref{ex:tawa}), it would prefer the prepositional interpretation (\ref{ex:tawa-prep}) when available.\footnote{The full implementation of the heuristic algorithm is found online at \url{https://github.com/cubedhuang/ilo-nasin-sin/blob/main/src/lib/parser.ts}.} This prioritization of interpretations reflects observed usage patterns in the Toki Pona community.

\subsection{Tagging parts of speech}

After sentences are represented hierarchically, a part-of-speech tagging module takes the parse structures as input and outputs a string of tokens tagged with parts of speech. The specific tags for Toki Pona are listed in Table~\ref{tab:partsofspeech}.

\begin{table}[hbt!]
  \caption{Token types and the contexts in which words are tagged by them\protect\footnotemark}
  \label{tab:partsofspeech}
  \begin{center}
    \begin{tabular}{p{0.20\linewidth} | p{0.70\linewidth}}
      \textbf{Token Type} & \textbf{Context} \\
      \hline
      \cat{Noun} & The first or only word in a subject, direct object, or prepositional object. \\
      \cat{Mod} & Modifiers, collectively adjectives and adverbs, that directly follow nouns and verbs. \\
      \cat{IVerb} & Intransitive verbs, typically the first word in the predicate when no direct object is supplied afterwards. \\
      \cat{TVerb} & Transitive verbs, typically the first word in the predicate when a direct object is supplied afterwards. \\
      \cat{Intj} & The first word in a phrase when it is used without a complete sentence. \\
      \cat{Prep} & Prepositions, specifically when interpreted prepositionally. \\
      \cat{Preverb} & Preverbs, lexical elements that precede a verb.\protect\footnotemark \\
      \cat{Part} & Grammatical particles and ordinal numbers. \\
    \end{tabular}
  \end{center}
\end{table}
\footnotetext[4]{A complete implementation of the tagging algorithm is found online at \url{https://github.com/cubedhuang/ilo-nasin-sin/blob/main/src/lib/tag.ts}.}
\footnotetext{The term `preverb' originates from \textcite{lang2014}'s description of Toki Pona's grammar. Preverbs are a closed class of words that go between the predicate marker and the main predicate, such as \foreign{wile} 'to want to (be),' \foreign{kama} 'to begin to; to become' (e.g. \foreign{jan li wile moku} 'the person wants to eat'). Like prepositions, they double as content words, so their interpretation as a preverb or as the main verb is decided by context.}

Tokens tagged as \cat{IVerb} or \cat{TVerb} are collectively referred to as \cat{Verb}, and the \cat{Noun}, \cat{Mod}, and \cat{Verb} tags are referred to as \cat{Content}.

\begin{figure}[hbt!]
  \centering
  \caption{Example of distribution of words across different content word classes for the year 2024 (\foreign{ma pona pi toki pona} corpus)}
  \includegraphics[width=0.75\textwidth]{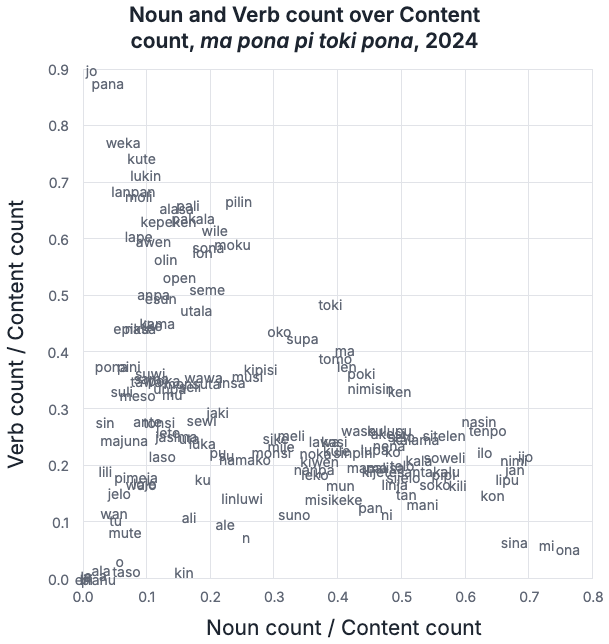}
  \label{fig:scatter1}
\end{figure}

When all tokens are tagged, the counts of each use of part of speech is aggregated by year, based on when the sentences were sent or published. For each year in each corpus, a distribution of words is created like in Figure~\ref{fig:scatter1}.\footnote{A full copy of the aggregated data as well as a set of visualization tools can be found online at \url{https://nasin.nimi.li/visualize}.}

\section{Results and discussion}

This study aimed to explore how content words' preferences for certain syntactic positions change over time and how the usage of content words in different syntactic positions differ between informal and formal contexts. The analysis revealed instances of diachronic variation in the use of several words and patterns that varied across two corpora.

Results will be presented in two sections: First, diachronic variation will be discussed, focusing on the changes in the use of body-part words as transitive verbs and the use of \foreign{pu} 'interacting with the official Toki Pona book' as a noun. Second, variation between the informal and formal corpora will be discussed, focusing on the use of interjections and the adoption of features across the two corpora.

\subsection{Diachronic variation}

The first research question aimed to examine how content words' preferences for certain syntactic positions change over time. In order to pursue the research questions, we chose words from the corpora that showed a high degree of change over time and whose change was referenced often in the Toki Pona community.

\subsubsection{Nouns as transitive verbs}

As discussed in Section~\ref{sec:fluidwordclasses}, because Toki Pona makes no syntactic distinction between content words, all content words can also become transitive verbs when followed by a direct object that is introduced with the transitive marker \foreign{e}. The body-part words \foreign{luka} 'hand, limb, branch' and \foreign{uta} 'mouth, lips' are nouns but can be used as transitive verbs meaning 'to touch by hand' and 'to touch by mouth' respectively, as shown in (\ref{ex:luka}) and (\ref{ex:uta}).

\pex\label{ex:luka}
\a \begingl
\glpreamble \foreign{luka} as a noun //
\gla waso li lukin e luka mi //
\glb bird \cat{pred} see \cat{tr} hand my //
\glft `The bird sees my hands.' //
\endgl
\a \begingl
\glpreamble \foreign{luka} as a transitive verb //
\gla jan li luka e waso //
\glb person \cat{pred} hand \cat{tr} bird //
\glft `The person touches the bird (with their hand).' //
\endgl
\xe

\pex~\label{ex:uta}
\a \begingl
\glpreamble \foreign{uta} as a noun //
\gla uta mi li pilin ike //
\glb mouth \cat{1.poss} \cat{pred} feel bad //
\glft `My mouth feels bad.' //
\endgl
\a \begingl
\glpreamble \foreign{uta} as a transitive verb //
\gla ona li uta e olin ona //
\glb 3 \cat{pred} mouth \cat{tr} love 3 //
\glft `They kissed the one they love.' //
\endgl
\xe

\begin{figure}[hbt!]
  \centering
  \caption{Change in usage of \foreign{luka} and \foreign{uta} as transitive verbs in \foreign{ma pona pi toki pona} from 2020 to 2024 as percentage of total content word usage.}
  \includegraphics[width=0.75\textwidth]{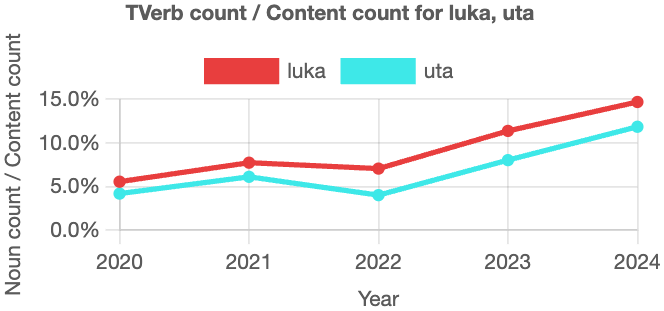}
  \label{fig:lukauta}
\end{figure}

These body-part words stand out because their usage in the informal corpus \foreign{ma pona pi toki pona} has drastically changed over time. As Figure~\ref{fig:lukauta} shows, the nouns \foreign{luka} and \foreign{uta} are increasingly used as transitive verbs in informal use of Toki Pona. This usage still falls within the grammar written in \textit{Toki Pona: The Language of Good}, as the meaning after conversion is a physical application of the noun to the direct object. However, only more recently has this usage proliferated in the community.

A possible explanation for this change is processing efficiency following \citeauthor{bybee2015}'s \parencite*[p.~1]{bybee2015} argumentation that ``mental processes ... are ... main causes of change'': body-part terms can be used as nouns to express similar ideas with the verb \foreign{pilin} 'to feel,' as in \foreign{jan li pilin e waso kepeken luka} 'the person feels the bird with their hand.' However, this construction is longer and more complex than simply using the body-part term as a transitive verb, as shown in (4b). Because the transitive construction is shorter while communicating the same information to the listener, it is likely to be preferred in informal contexts where brevity is valued.

\subsubsection{Adjectives as nouns}

The adjective \foreign{pu} was defined in 2014 to mean ``interacting with the official Toki Pona book'' \parencite{danielson2024,lang2014}. The typical way to refer to the book \textit{Toki Pona: The Language of Good} in the language is with the phrase \foreign{lipu pu} 'the book that is the official Toki Pona book.' The word \foreign{pu} can be used in a variety of contexts, such as the sentence \foreign{ona li pu} 'they are reading the official Toki Pona book' and the phrase \foreign{nasin pu} 'the ways of the official Toki Pona book.'

\begin{figure}[hbt!]
  \centering
  \caption{Change in the use of \foreign{pu} as a modifier and noun in \foreign{ma pona pi toki pona}.}
  \includegraphics[width=0.75\textwidth]{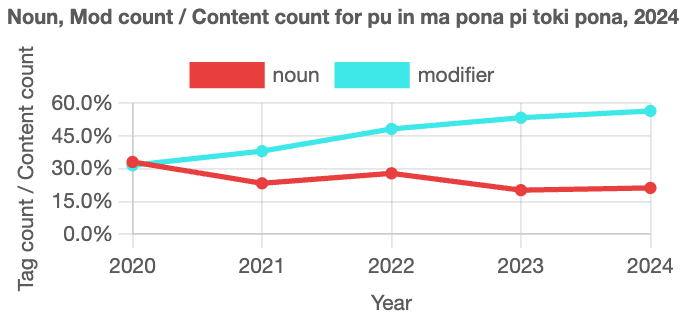}
  \label{fig:pumod}
\end{figure}

Many speakers instead opt to refer to the book simply with \foreign{pu} as a standalone noun without \foreign{lipu} 'document' preceding it. However, more recently, this noun usage referring to the book has decreased, with speakers switching back to referring to the book using the full phrase \foreign{lipu pu}, as seen in Figure~\ref{fig:pumod}.

This is likely due to pressure to conform with existing name constructions in the language, applying known patterns to new contexts and thereby reinforcing these patterns \parencite[cf.][p.~10]{bybee2015}. In Toki Pona, names must be adjectives and cannot be nouns: a person named \foreign{Tani} is referred to as \foreign{jan Tani} 'Tani the person,' and a country named \foreign{Kanata} is referred to as \foreign{ma Kanata} 'Kanata the place.' This construction is similar to the typical way of referring to the book, \foreign{lipu pu}. Toki Pona speakers may encounter the construction \foreign{lipu pu} enough that they analyze \foreign{pu} more as a proper name attached to \foreign{lipu} rather than a typical content word. As such, Toki Pona users become more likely to match the form of these other constructions.

\subsection{Variation across corpora}

The second research question aimed to examine how the usage of words in different syntactic positions vary between informal and formal contexts.

\subsubsection{Use of interjections}

In conversational text, speakers will often use words as standalone interjections rather than full sentences. For example, saying \foreign{pona} 'good' alone is often a sign of acknowledgement, like English 'okay.' Interjections are also used to answer polar questions by either repeating the verb or its negation.

\begin{figure}[hbt!]
  \centering
  \caption{Usage of \foreign{pona} as a standalone interjection between the two corpora.}
  \includegraphics[width=0.75\textwidth]{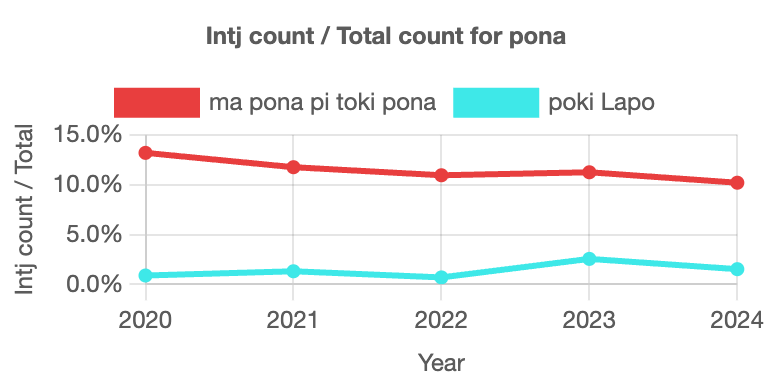}
  \label{fig:ponaintj}
\end{figure}

As seen in Figure~\ref{fig:ponaintj}, there is a consistent difference in the use of \foreign{pona} as an interjection between the corpora, and the ratio remains relatively stable over time.

\begin{figure}[hbt!]
  \centering
  \caption{Usage of \foreign{wawa} as a standalone interjection between the two corpora.}
  \includegraphics[width=0.75\textwidth]{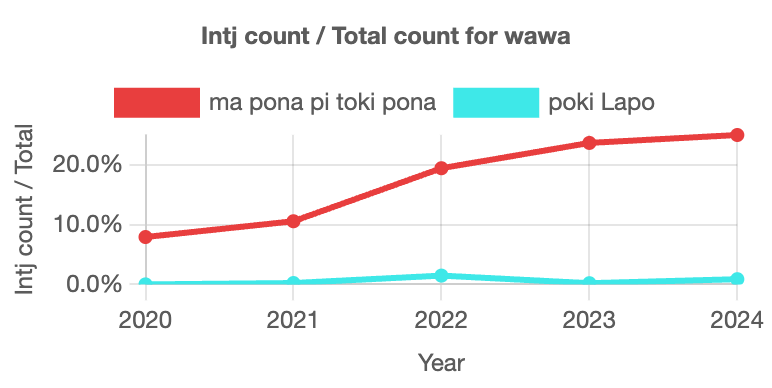}
  \label{fig:wawaintj}
\end{figure}

The usage of \foreign{wawa} 'strong, powerful' as an interjection has drastically increased in conversational data as seen in Figure~\ref{fig:wawaintj}. Despite this increase, an increase in usage has not been observed in the formal corpus.

This pattern can be attributed to several factors. First, interjections may inherently appear less in non-conversational data because simpler statements like acknowledgements are less necessary. Second, formal writing and longer prose tends to be more impersonal and disembodied \parencite{mcculloch2020}. In such contexts, interjections that express emotions and exclamations are less likely to be included.

\subsubsection{Adoption of features}

\begin{figure}[hbt!]
  \centering
  \caption{Usage of \foreign{luka} as a transitive verb between the two corpora.}
  \includegraphics[width=0.75\textwidth]{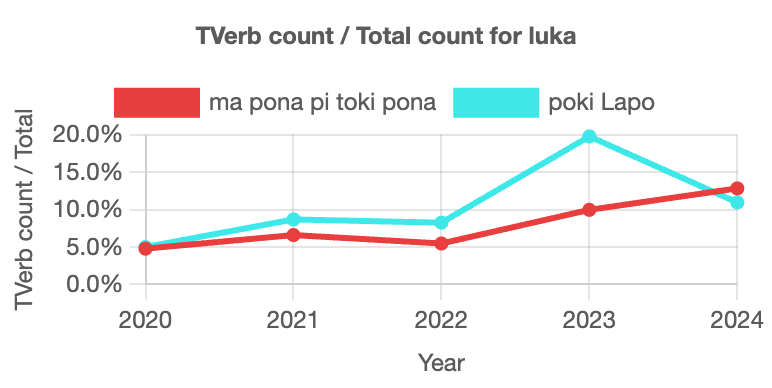}
  \label{fig:lukatwo}
\end{figure}

Because language change typically occurs slowly, adoption of linguistic innovations was expected to be delayed in more formally written texts where a more `traditional' grammar might be preferred in order to prioritize accessibility to a wider audience. However, the adoption of features is not delayed in the formal corpus. Instances of diachronic variation observed in the informal corpus were observed in the formal corpus at the same time, as in Figure~\ref{fig:lukatwo}.

One explanation for the lack of an apparent difference between the two corpora is the homogeneity of the Toki Pona community. A significant majority of authors with published works in Toki Pona are members of the \foreign{ma pona pi toki pona} Discord or are heavily connected to it; the language has a tightly knit community of speakers. Additionally, though increases in body parts as transitive verbs and \foreign{pu} as a modifier are observed, these changes still fit within the grammar and derivation patterns described in \textit{Toki Pona: The Language of Good}, so writers may feel that these usages are still understandable to a reading audience. Continued repetition then may have led to the adoption of features in Toki Pona in general, formal and informal uses \parencite[cf.]{bybee2015}. Adoption of features may also be understood as an act of identity \parencite[p.~193]{labov2010}. In order to identify with the Toki Pona community, one may choose expressions that reflect its communicative convention, reinforcing the feature in turn.

\section{Conclusion}

This study used a computational and corpus-based approach to examine language variation over time and across genres of the constructed language Toki Pona. The results of the parsing and part-of-speech tagging analysis revealed that the word classes of certain lexical items are associated with change over time, indicating that changes noticed by speakers of Toki Pona community were true. For instance, nouns depicting body-parts are increasingly used as transitive verbs, and proper nouns may be used more as adjectives.

Although the study contributes valuable insights to the field of language change and variation, the results should be taken with care as the study has a few limitations. The available time period of Toki Pona is relatively short (2020-2025) compared to studies of natural languages. Furthermore, because both corpora contained written data, generalizations about spoken variations cannot be made.

Future research could determine whether semantic features found in natural languages extend to Toki Pona. For instance, \textcite{glass} found that in English, pragmatic factors make subjective adjectives more likely predicative and objective adjectives more likely attributive, and \textcite{dyer2020predictingcrosslinguisticadjectiveorder} found that the subjectivity rating of adjectives in a language is a strong predictor of distance from the noun. These methods could be replicated in Toki Pona by scoring subjectivity for Toki Pona's content words and evaluating the prediction power in the same corpora.

To conclude, this study has shown language change and variation in a constructed language, Toki Pona. The linguistic innovations all occur within the existing grammatical framework rather than expanding on it. Toki Pona's fluid word class system allows for unique usage without losing intelligibility. Thus, constructed languages are influenced by cognitive and sociolinguistic factors such as processing efficiency, conforming of known structural patterns to new contexts, and enactment of identity as member of the Toki Pona community just like natural languages \parencite[cf.][]{bybee2015,labov2010}. The findings suggest that language communities naturally balance innovation and stability, maintaining mutual intelligibility while finding new ways to communicate. Thus, innovation within constraints may be a fundamental property of language use.

\printbibliography

@online{leipzigglossingrules,
  title = {{Leipzig Glossing Rules}},
  url = {https://www.eva.mpg.de/lingua/resources/glossing-rules.php},
  author = {Bernard Comrie and Martin Haspelmath and Balthasar Bickel},
  publisher = {Department of Linguistics of the Max Planck Institute for Evolutionary Anthropology & the Department of Linguistics of the University of Leipzig},
  year = {2024},
  month = {01},
  day = {01}
}

@book{peterson2015,
  address={New York},
  title={{The Art of Language Invention: From Horse-Lords to Dark Elves to Sand Worms, the Words Behind World-Building}},
  ISBN={9780143126461},
  publisher={Penguin Books},
  author={Peterson, David J.},
  year={2015},
  language={eng}
}

@book{bybee2015,
  place={Cambridge},
  series={Cambridge Textbooks in Linguistics},
  title={{Language Change}},
  publisher={Cambridge University Press},
  author={Bybee, Joan},
  year={2015},
  collection={Cambridge Textbooks in Linguistics}
}

@book{labov2010,
  author = {Labov, William},
  publisher = {{John Wiley \& Sons, Ltd}},
  isbn = {9781444327496},
  title = {Principles of Linguistic Change},
  doi = {10.1002/9781444327496.ch9},
  year = {2010},
}

@online{dyer2020predictingcrosslinguisticadjectiveorder,
    title={Predicting cross-linguistic adjective order with information gain}, 
    author={William Dyer and Richard Futrell and Zoey Liu and Gregory Scontras},
    year={2020},
    eprint={2012.15263},
    archivePrefix={arXiv},
    primaryClass={cs.CL},
    url={https://arxiv.org/abs/2012.15263}, 
}

@online{danielson2025,
  author = "Danielson, III, Gregory",
  title = {{sona-toki}},
  day = {14},
  month = apr,
  year = {2025},
  publisher = {GitHub},
  note = {GitHub repository},
  url = {https://github.com/gregdan3/sona-toki},
}

@online{danielson2024,
  author = "Danielson, III, Gregory",
  title = "When was {pu} added to Toki Pona?",
  year = 2024,
  month = mar,
  day = 1,
  url = {https://web.archive.org/web/20240301100156/https://mun.la/lipu/pu.html},
  urldate = {2024-03-01}
}

@online{lapo2025,
  author = "{kala Asi} and {ijo vivi} and {jan Juwan} and {jan Kita}",
  title = {{poki Lapo}},
  day = {12},
  month = apr,
  year = {2025},
  publisher = {GitHub},
  note = {GitHub repository},
  url = {https://github.com/kulupu-lapo/poki},
}

@online{mapona2025,
  title = "{ma pona pi toki pona}",
  url = {https://discord.gg/mapona},
  year = {2025},
  month = jan,
  day = {9},
  note = {Discord server},
  howpublished = {Online},
}

@online{nearley,
    author = "Kartik Chandra and Tim Radvan",
    title  = "{nearley}: a parsing toolkit for {JavaScript}",
    year   = {2020},
    month  = jun,
    day   =  {17},
    doi    = {10.5281/zenodo.3897993},
    url    = {https://github.com/kach/nearley}
}

@article{chomsky1956,
  author={Chomsky, N.},
  journal={IRE Transactions on Information Theory}, 
  title={Three models for the description of language}, 
  year={1956},
  volume={2},
  number={3},
  pages={113-124},
  doi={10.1109/TIT.1956.1056813}}

@online{muni2024,
  author = "Danielson, III, Gregory",
  title = {{ilo Muni}},
  year = 2024,
  month = aug,
  day = 10,
  url = {https://ilo.muni.la},
}

@online{meulen2021,
  author = {Meulen, Spencer van der},
  title = {{Request for New Language Code Element in ISO 639-3}},
  year = 2021,
  month = oct,
  day = 31,
  url = {https://iso639-3.sil.org/sites/iso639-3/files/change_requests/2021/2021-043_tok.pdf}
}

@article{glass,
  author = {Glass, Lelia},
  year = {2024},
  month = {08},
  pages = {},
  title = {The red dress is cute: Why subjective adjectives are more often predicative},
  journal = {Corpus Linguistics and Lingustic Theory},
  doi = {10.1515/cllt-2024-0044}
}

@article{earley1970,
  author = {Earley, Jay},
  title = {An efficient context-free parsing algorithm},
  year = {1970},
  issue_date = {Feb 1970},
  publisher = {Association for Computing Machinery},
  address = {New York, NY, USA},
  volume = {13},
  number = {2},
  issn = {0001-0782},
  url = {https://doi.org/10.1145/362007.362035},
  doi = {10.1145/362007.362035},
  journal = {Commun. ACM},
  month = feb,
  pages = {94–102},
  numpages = {9}
}

@book{lang2014,
  place={United States},
  title={{Toki Pona: The Language of Good}},
  publisher={Tawhid},
  author={Lang, Sonja},
  year={2014}
}

@book{mcculloch2020,
  title={{Because Internet: Understanding how Language is Changing}},
  author={McCulloch, G.},
  isbn={9781529112825},
  year={2020},
  publisher={Vintage}
}
\end{document}